\RequirePackage{amsmath}
\documentclass{bjmc}
\usepackage{multibib}
\usepackage{graphicx}
\usepackage{tabularx}
\usepackage{soul}
\usepackage{times}
\usepackage{url}
\usepackage{latexsym}
\usepackage{layout}
\usepackage{float}
\usepackage{placeins}
\usepackage{wrapfig}
\usepackage{afterpage}
\usepackage[normalem]{ulem}
\usepackage{color}
\usepackage{nicefrac}
\usepackage{multirow}
\usepackage{amsmath}
\usepackage{natbib}
\usepackage{pgfplots}
\pgfplotsset{compat=1.14}
\usepackage{tikzscale}
\usepackage{anyfontsize}
\usepackage[utf8]{inputenc}
\usepackage{epstopdf}
\usepackage{hyperref}
\usepackage{xstring}
\usepackage{ifdraft}
\usepackage{courier}

\let\subparagraph\paragraph
\usepackage[compact]{titlesec}
\titlelabel{\thetitle.\quad}

\doi{https://doi.org/10.22364/bjmc.2022.10.3.11}
\proceedings{10}{3}{381}
\begin{document}
\mainmatter  
\title{How Masterly Are People at Playing with Their Vocabulary?}
\titlerunning{How Masterly Are People at Playing with  Their Vocabulary?}

\author{Matīss RIKTERS\inst{1}, Sanita REINSONE\inst{2}}
\authorrunning{Rikters and Reinsone}

\setcounter{page}{382}
\institute{University of Tartu, Estonia \\
    \and Institute of Literature, Folklore and Art, University of Latvia \\
    \texttt{matiss.rikters@ut.ee, sanita.reinsone@lulfmi.lv}}
\maketitle
\begin{abstract}
    In this paper, we describe adaptation of a simple word guessing game that occupied the hearts and minds of people around the world. There are versions for all three Baltic countries and even several versions of each. We specifically pay attention to the Latvian version and look into how people form their guesses given any already uncovered hints. The paper analyses guess patterns, easy and difficult word characteristics, and player behaviour and response. 
\end{abstract}

\section{Introduction}
\label{intro}

Word guessing games are a phenomenon that represents the use of language in unusual socio-cultural contexts. Depending on the rules of a  game, the meaning of a word can be completely irrelevant, whereas the structural elements of a word, such as its length and the composition of its letters, may play an important role. Despite words being used as game attributes without the need to know their meaning and context of use, it can be argued that such games contribute to vocabulary mastery and general language training. 

One of the first computer games created in Latvia in the mid-1990s was the word scoring game Lingo. Inspired by a popular TV show, it was created by the language technology company Tilde. The game required guessing a five-letter word, the first letter of which was known to a player. As one of the few games available on almost all computers in Latvia at the time, it became very popular among players of all ages. The game's word corpus contained 999 words \citep{lingo-delfi}. The game was required to be installed on a computer and could be played offline without restriction or any limit.

Almost thirty years later, in October 2021, an online word-puzzle game Wordle\footnote{\url{https://www.nytimes.com/games/wordle/index.html}} was invented by a software engineer Josh Wardle. In a relatively short time, it became globally popular, attracting more and more new players and spawning new language versions around the world. The principle of the game is fairly simple and similar to Lingo – a player is given six attempts to guess a five-letter word. After each guess, the letters are coloured in three colours (grey, orange, green), giving the player a hint on how to continue guessing the hidden word. Unlike other word games, the specifics of Wordle are that only one word can be guessed per day, which is the same for all players. For it to work, players must be disciplined not to reveal the word of the day to others. 

However, to share results instantly without spoiling the enjoyment of the game for others, Wordle offers to create an abstract figure made up of emoji library squares in three colours. It contains a geometric pattern that shows the progress and result of a guess without revealing the word behind it. This figure that players share on social networks, is the most important representational attribute of the game, also acting in a symbolic way as a communication and interaction element within the community.

Although the original Wordle is in English, enthusiast developed open-source code behind the game enables it to be adapted for other languages.
GitHub hosted 'Wordles of the World'\footnote{\url{https://rwmpelstilzchen.gitlab.io/wordles}} list contains links to Wordle games in more than 90 languages. For example, at least three versions of Wordle game have been currently developed for each – Estonian, Latvian and Lithuanian: 
\begin{itemize}
    \item Estonian versions:
    \begin{itemize}
      \item \url{https://uudis.net/wordle}
      \item \url{https://sonuk.subscribe.ee}
      \item \url{https://sonar.ajad.ee}
    \end{itemize}
    
    \item Latvian versions:
    \begin{itemize}
      \item \url{https://wordle.lielakeda.lv}
      \item \url{https://ralfulis.vercel.app}
      \item \url{https://vardulis.lv}
    \end{itemize}
    
    \item Lithuanian versions:
    \begin{itemize}
      \item \url{https://jakut.is/vordl}
      \item \url{https://dienos-zodis.lt}
      \item \url{https://wordle.dario.cat}
    \end{itemize}
\end{itemize}

While the Wordle developer admitted that the game is most appreciated precisely because of the fun it brings \citep{nyt-wordle}, Wordle users and re-designers have managed to add additional value to the game showing potential for promoting learning – it is being used in education for new language acquisition \citep{Brown2022,wordle-classroom}, as well as to revitalise endangered languages \citep{wordle-lang,wordle-cbc}.

\section{Game Construction}
\label{sec:tool}

Shortly after the swift rise in popularity of the original Wordle game several versions of its reconstruction started popping up on GitHub. Of these the most popular became React Wordle \footnote{\url{https://github.com/cwackerfuss/react-wordle}}, which so far has over 1,700 forks and over 2,200 stars, and has been used as a base to create Wordle versions in 43 different languages (even Latin and Cornish), 32 thematic versions (such as birds, super heroes, airport codes), and even 20 mathematics, science, technology oriented ones (for example, gene symbols, JavaScript, prime numbers). The base code, which was made using React, Typescript, and Tailwind libraries, has been developed for easy adaption to new languages or themes. For example, to have a personal list of daily words and valid guesses only two files need to be updated, and to adapt the code to a new language 7 to 9 other files need changes, for which detailed instructions have are provided in the GitHub  repository.

\subsection{Adapting Wordle into Latvian and Audience Involvement}

The first Latvian version of Wordle was created in mid January 2022. The game was named `Vārdulis' – deriving its title from Latvian `vārds' (word), but keeping a sonic resemblance to original title. Giving the game a unique, Latvian-specific name was a successful choice, as it was easy to find Vārdulis mentions on social media from day one of the game's launch. This, in turn, is essential for communication between players. 

Even though Wordle is meant to be played in a single-player manor, an essential part of the game is sharing the result, i.e. game’s auto-generated grids of emoji squares, and discussing the word of the day without revealing it on social media, such as Twitter\footnote{\url{https://twitter.com/search?q=vardulis}}, or internal communication tools, such as WhatsApp. The impact of social media on the popularity of the game is significant. Sharing a score is often a conversation starter with other, previously unknown players, it is also a micro-competition to compare who has the better score and more successful choice of words. The words of the day are discussed and evaluated mostly in terms of their game-specific difficulty. 

When developing the Latvian version, the decision was made to also include person names and various inflections of the words instead of plain singular nominative forms of nouns, incorporating words that have four letters in the nominative case, but five when conjugated (e.g., flower: nom. `puķe' – gen. `puķes'), thus making Vārdulis much more of a challenge than its English counterpart. However, a decision to include such words was reached in order to highlight the diversity of the language and have a more abundant set of data for subsequent play analysis. In addition, to include the possibility to learn more about the meaning of words, a link to the word entry in the online dictionary and thesaurus\footnote{\url{https://tezaurs.lv}} developed by the Institute of Mathematics and Informatics of the University of Latvia was included in the window that pops up when the game is finished.

In the public discussions on Twitter which is the most common public space for Vārdulis players to meet, it can be seen that the most topical issue regarding Vārdulis is the extended dictionary, i.e. the inclusion of inflected forms in the game. The criticism was particularly strong in the first months of the game. Players complained that the game's rules thus are not fair and that they should stick to the rules of the original version, that the Latvian version is too complicated, that there are too many conjugations in Latvian to win the game in six attempts. It is also joked that the title of the game should rather be ``guess the correct conjugation".\footnote{\url{https://twitter.com/DavisVilums/status/1489537145836609537}} Over time, the criticism decreased, players accepted the rules and the vocabulary used by players increased. 

Vārdulis, just like its original Wordle is limited to one game per day. The average game sessions per day from January 28 to April 14, 2022 is 935, however it took around 2 weeks for popularity to rise from a few hundred plays to around a thousand per day.

By exploring the user statistics of tezaurs.lv in Google Analytics,\footnote{For the purposes of the research, the Institute of Mathematics and Informatics of the University of Latvia granted access to the Google Analytics account of tezaurs.lv.} it can be seen that the daily word is one of the most frequent searches in the database on a given day. On average, 5.7\% of players navigate to the thesaurus to explore a particular word. 

Exploring which words are most frequently consulted in the tezaurs.lv, two tendencies can be observed: first, less known or unusual word, for example, `adobe' (meaning air-dried clay brick) that many of the players have never heard in Latvian was searched for on tezaurs.lv by 61.97\% of players. Secondly, words that were difficult to guess or that a large number of players failed to guess at all. For example, 40.81\% of players failed to guess quite common word `šuves' (stitches), accordingly, on the given day, 21.14\% of players searched for this word on the tezaurs.lv.

Overall, it can be concluded that the linking of tezaurs.lv with Vārdulis is successful and serves its purpose well, but it could also be used in a more targeted way by regularly including less known and used words in the list of daily words, which would provide additional opportunities for mastering vocabulary of players. However, as the game is to some extent competitive and players aim to complete the game in as few attempts as possible, players' frustration and public complaining could be expected.

\subsection{Word List Generation}
\label{sec:scores}

There are two word lists necessary to play the game – a list of daily guesses (main list) and a list of all valid guesses (secondary list). Construction of both lists was performed semi-automatically. First, we acquired all monolingual Latvian corpora from Opus \citep{TIEDEMANN12.463}, tokenised the data, filtered out only tokens consisting of 5 characters, and finally removed any tokens which had any character outside the 33 character Latvian alphabet. To make the game reasonably challenging, we ordered the remaining tokens by frequency of occurrence in the corpora and chose the 1,500 most frequent words for the main list and everything else for the secondary list. 

To maintain purely words in the Latvian language, we cross-referenced the list with the Lexical Database for Latvian \citep{spektors-etal-2016-tezaurs} and manually reviewed each word. After this, 1,430 words remained in the main list while some very frequent foreign words such as ``China" or ``Apple" were removed. We selected the further 15,000 words from the list ordered by frequency for the secondary list, also cross-referencing with the Lexical Database, but without manually verifying.

The secondary list, however, was still at times falling short of it's objective by failing to recognise perfectly valid Latvian words in specific inflections which may not have necessarily been among the 16,500 most frequent five-character words in the corpora. To improve the list, we once again turned to the Lexical Database and selected all words in lengths of 3 to 8 characters, automatically inflected them to all possible word forms using an inflection generator \citep{nikiforovs2011}, and filtered the results down to inflections of the words spanning exactly 5 characters. While still not fully exhausted, the secondary list grew to 22,341 words.

\section{Play Analysis}
\label{sec:debugging}

The design of our version of the game includes logging the array of guesses for each session played until the end (either correct guess or failed after six attempts). In this section we analyse game data of 77 daily words collected between January 28th and April 14th of 2022. 

\begin{table}[t]
\centering
\caption{Top 15 guesses at each turn. Words that were the actual answers within these days are marked in bold. English translations of the words can be found in \hyperref[sec:top-15-en]{Appendix B}.}
\begin{tabular}{|l|r|l|r|l|r|l|r|l|r|l|r|}
\hline
\multicolumn{1}{|c|}{\textbf{G1}} & \multicolumn{1}{c|}{\textbf{$\sum$}} & \multicolumn{1}{c|}{\textbf{G2}} & \multicolumn{1}{c|}{\textbf{$\sum$}} & \multicolumn{1}{c|}{\textbf{G3}} & \multicolumn{1}{c|}{\textbf{$\sum$}} & \multicolumn{1}{c|}{\textbf{G4}} & \multicolumn{1}{c|}{\textbf{$\sum$}} & \multicolumn{1}{c|}{\textbf{G5}} & \multicolumn{1}{c|}{\textbf{$\sum$}} & \multicolumn{1}{c|}{\textbf{G6}} & \multicolumn{1}{c|}{\textbf{$\sum$}} \\ \hline
SAULE & 5341 & \textbf{LAIKS} & 412 & \textbf{LAIKS} & 433 & \textbf{TĒRPU} & 432 & \textbf{TĪRĪT} & 382 & \textbf{FLĪŽU} & 345 \\ \hline
SIENA & 3179 & SAULE & 355 & \textbf{TIESA} & 334 & \textbf{TĪRĪT} & 382 & \textbf{DARĀT} & 364 & \textbf{RAIŅA} & 296 \\ \hline
\textbf{TIESA} & 1579 & DIENA & 337 & \textbf{TAUKI} & 295 & \textbf{PUSEI} & 382 & \textbf{ILGĀK} & 359 & \textbf{BAUDU} & 295 \\ \hline
DIENA & 1476 & LIEPA & 290 & \textbf{LIETU} & 273 & \textbf{VĒLĀK} & 380 & \textbf{GROZĀ} & 350 & \textbf{MAIGI} & 289 \\ \hline
LAIME & 1449 & KAĶIS & 284 & \textbf{PUSEI} & 271 & \textbf{GARĀM} & 364 & \textbf{SAVĀM} & 340 & \textbf{BIEŽA} & 278 \\ \hline
PIENS & 1237 & PIENS & 266 & \textbf{LAIKU} & 247 & \textbf{LAIKS} & 354 & \textbf{TĒRPU} & 337 & \textbf{CELTA} & 258 \\ \hline
MAIZE & 1217 & ĀBOLS & 262 & \textbf{PRECE} & 230 & \textbf{KURSĀ} & 353 & \textbf{VĒRTS} & 334 & \textbf{SAVĀM} & 250 \\ \hline
LIEPA & 1159 & SAULĒ & 207 & \textbf{DIEVS} & 225 & \textbf{KRĀSU} & 340 & \textbf{ZEMĒM} & 327 & \textbf{JĀŅEM} & 245 \\ \hline
SAITE & 958 & SIENA & 205 & \textbf{LIKTS} & 224 & \textbf{LIKTS} & 339 & \textbf{IELEJ} & 324 & \textbf{PLAŠA} & 243 \\ \hline
KASTE & 952 & LIETA & 204 & \textbf{PUSES} & 214 & \textbf{PRECE} & 338 & \textbf{GARĀM} & 322 & \textbf{LABAS} & 238 \\ \hline
ĀBOLS & 950 & MAIZE & 203 & \textbf{TIRGU} & 214 & \textbf{GALDU} & 335 & \textbf{LABAS} & 321 & \textbf{ZEMĒM} & 237 \\ \hline
KAĶIS & 869 & RIEPA & 192 & \textbf{TĒRPU} & 206 & \textbf{DIEVS} & 332 & \textbf{VĒLĀK} & 321 & \textbf{ZINOT} & 236 \\ \hline
IELAS & 676 & LAIME & 188 & \textbf{MIERU} & 201 & \textbf{BLOKU} & 331 & \textbf{TĀPAT} & 317 & \textbf{IELEJ} & 234 \\ \hline
SKOLA & 673 & \textbf{TAUKI} & 175 & \textbf{REIZĒ} & 200 & \textbf{TIRGU} & 328 & \textbf{JĀŅEM} & 317 & \textbf{KAKLU} & 226 \\ \hline
\end{tabular}
\label{tab:guess-table}
\end{table}

Table \ref{tab:easy-words} shows the top 10 most difficult words to guess ordered by the amount of plays where the player was unable to guess the word after six guesses, and top 10 easiest words ordered where only very few players were unable to find the correct word while most were successful after the third or fourth guess. Here it is visible that a good deal of the easy words are nouns in singular nominative form, most of them do not contain diacritics, and have almost no repetition of characters within the word. On the other hand, most of the difficult words contain at least one or two diacritics, have repeating characters within the word, and none of the words are in singular nominative nouns.

\begin{figure}[h]
  \includegraphics[width=\linewidth]{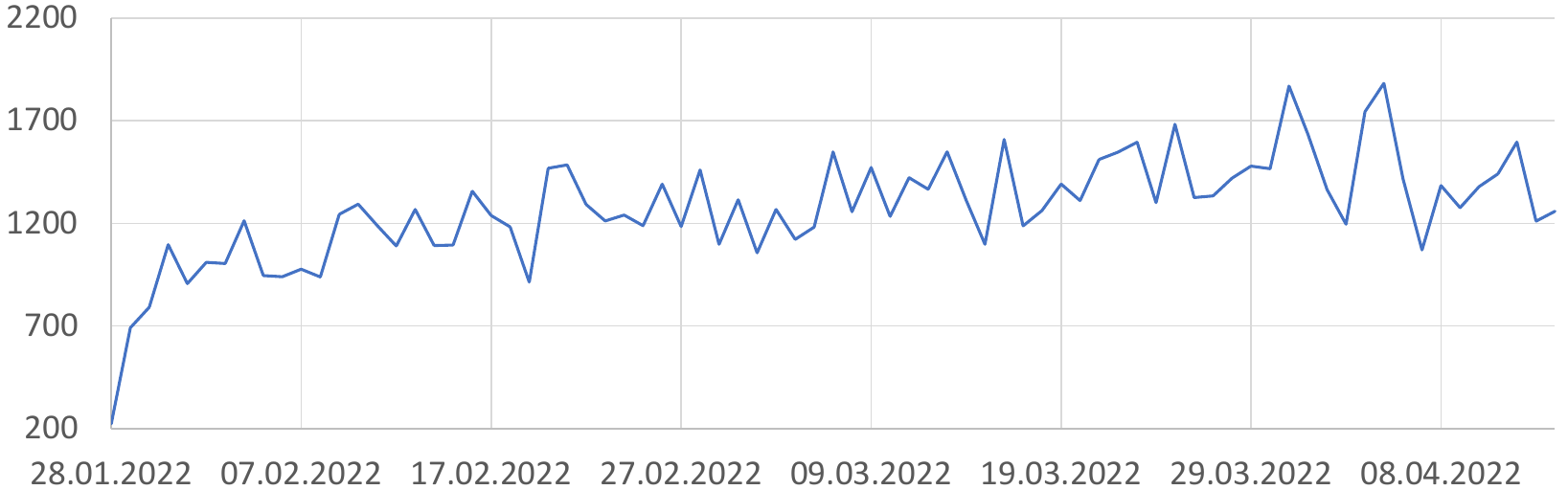}
  \caption{Change of unique word forms used for guessing over time.}
  \label{fig:uniq-wf}
\end{figure}

The total number of tokens used by Vārdulis players in 77 days is 12,705. As it can be seen in Figure \ref{fig:uniq-wf}, the vocabulary used by players tends to expand. Table \ref{tab:guess-table} shows the most popular word choices at each stage of the game. All words in columns G3-G6 have been the correct word of the day at some point. From the opening guess column G1 we clearly see that most players start with a singular nominative noun without diacritics, and with no overlapping characters within the word to make use of uncovering hints for future guesses. An interesting observation in Table \ref{tab:guess-table} is that the most popular opening word by far is ``Saule" (the Sun), followed by ``siena" (wall), and ``tiesa" (court or truth).

\begin{table}[h]
  \caption{Easiest and most difficult words to guess. Row C indicates the number of occurrences of the word in the specific form in the corpus, rows G1-G6 represent guesses, and row X represents failed games after 6 guesses. English translations of the words can be found in \hyperref[sec:easy-diffi-en]{Appendix A}.}
  \includegraphics[width=\linewidth]{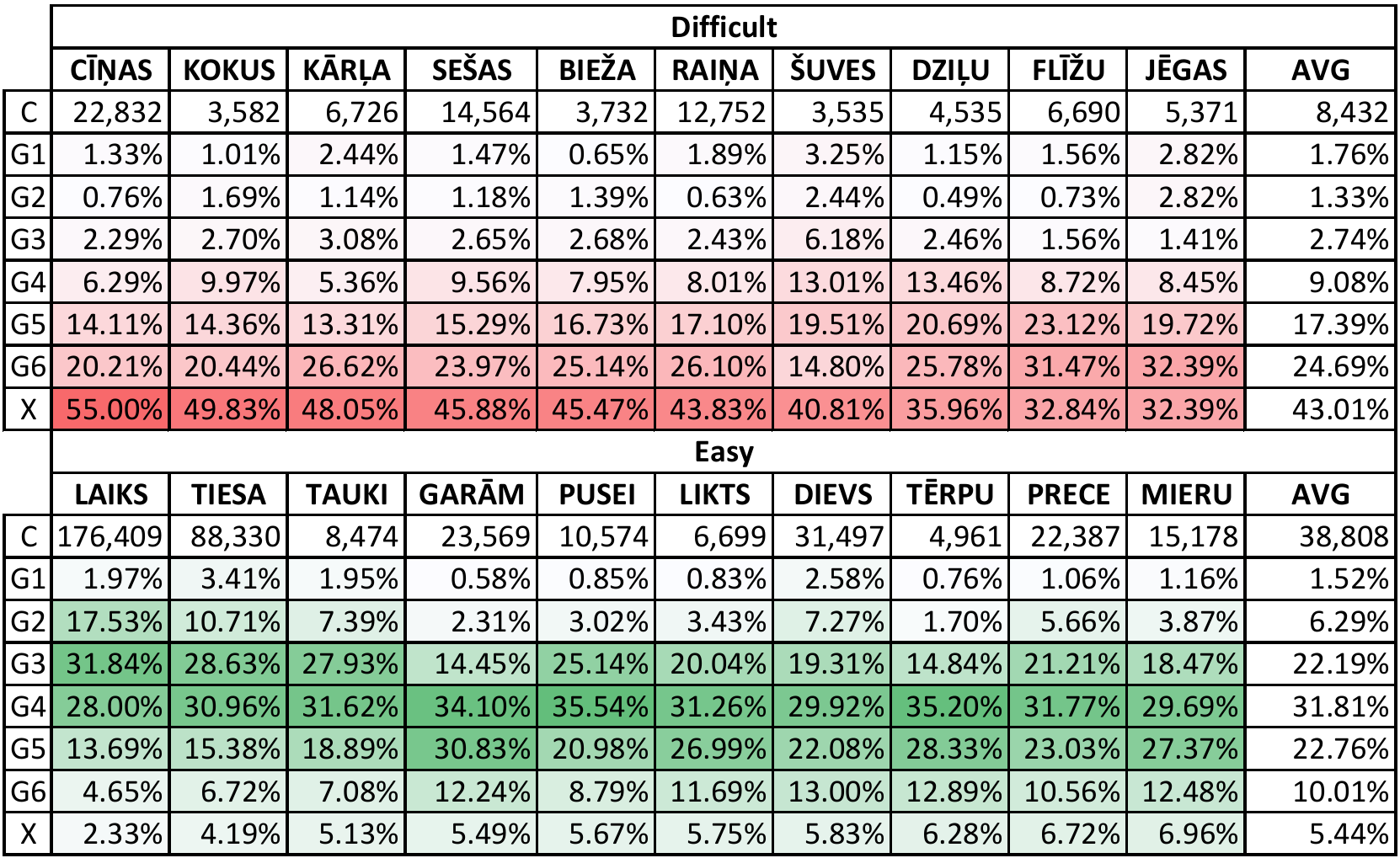}
  \label{tab:easy-words}
\end{table}

\begin{figure}[h]
    \centering
    \includegraphics[width=9.5cm]{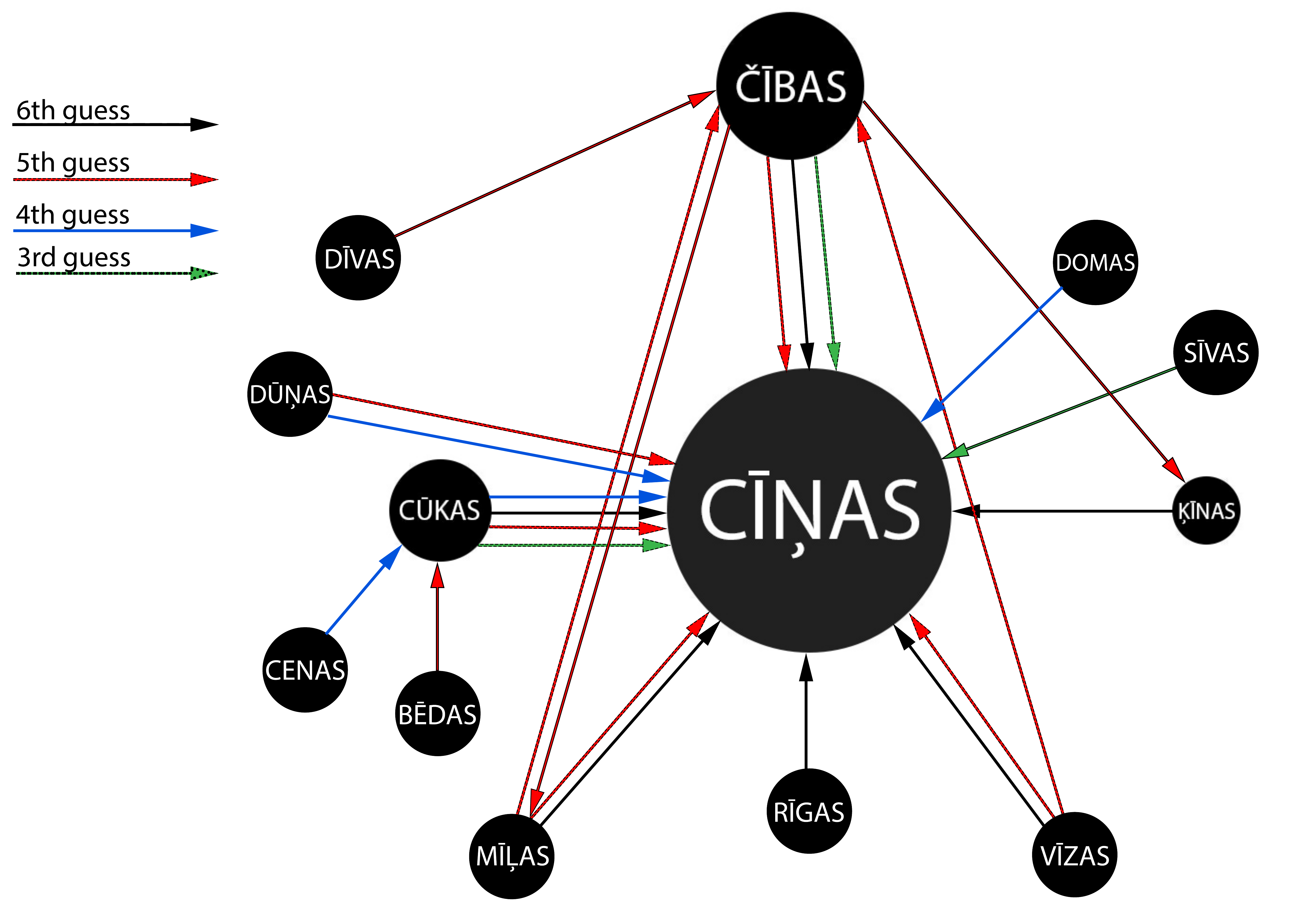}
    \caption{Paths of previous guesses that lead players to the correct answer for the most difficult word of the day so far – ``CĪŅAS".}
    \label{fig:hard-word}
\end{figure}

We look in detail at the most challenging word so far in the game and depict most common guess paths taken by players in Figure \ref{fig:hard-word}. The different arrows show at which of the six attempts to guess players were at. It is visible here that the vast majority of guesses at the last stages had already uncovered the ending of the correct answer ``AS", and some had other critical characters uncovered like ``Ī", ``C" or ``Ņ".

\section{Conclusion}
\label{sec:conclusion}

In this paper, we provided insight in a brief linguistic exercise that has become a fun pastime for a few minutes each day for many players around the world. The creation of a near complete Latvian version of the game is described with further hints on how to make it more or less challenging and the possibility of enriching the vocabulary by linking the game to an online thesaurus is examined. While providing a glimpse into the public perception of the Latvian version of the game, we also dive deep in analysing how the Latvian word game has been played over the first two and a half months, looking at players' strategies, easier and more difficult words to guess.

In future work, we plan to automatically analyse each daily word morphologically and attempt to predict the difficulty level or even guess the distribution based on a machine learning model.

\section{Acknowledgements}
\label{sec:acknowledgments}

This work has received funding from the “European Social Fund via IT Academy programme” and the project "Research on Modern Latvian Language and Development of Language Technology” (No. VPP-LETONIKA-2021/1-0006).


\bibliographystyle{apalike}
\bibliography{bibliography}

\FloatBarrier

\appendix

\section*{Appendix A. Translations of Easy and Difficult Words}
\label{sec:easy-diffi-en}

Table \ref{tab:easy-diffi-en} shows the English translations and accompanying the part-of-speech tags \citep{oxford} of the easiest and most difficult words to guess (from Table \ref{tab:easy-words}).

\begin{table}[h]
\centering
\caption{Translations of the easiest and most difficult words to guess with accompanying the part-of-speech tags.}
\begin{tabular}{|c|c|c|c|c|c|c|}
\hline
\multirow{4}{*}{\textbf{Difficult}} & \textbf{Word} & Battle & Trees & Carl's & Six & Frequent \\ 
 & \textbf{Tag} & N nom pl & N acc pl & N gen sg & Num nom pl & Adj nom sg \\ \cline{2-7} 
 & \textbf{Word} & Rainis' & Stiches & Deep & Tiles & Sense \\ 
 & \textbf{Tag} & N gen sg & N nom pl & Adj acc sg & N acc pl & N gen sg \\ \hline
\multirow{4}{*}{\textbf{Easy}} & \textbf{Word} & Time & Court & Fat & Past & Half \\  
 & \textbf{Tag} & N nom sg & N nom sg & N nom pl & Adv & N dat sg \\ \cline{2-7} 
 & \textbf{Word} & Put & God & Outfit & Product & Peace \\ 
 & \textbf{Tag} & V ptcp pst m & N nom sg & N acc sg & N nom sg & N acc sg \\ \hline
\end{tabular}
\label{tab:easy-diffi-en}
\end{table}

\FloatBarrier
\newpage
\section*{Appendix B. English Translations of Top 15 Guesses}
\label{sec:top-15-en}
English translations and accompanying the part-of-speech tags \citep{oxford} of the top 15 guesses at each turn (from Table \ref{tab:guess-table}) are shown in Table \ref{tab:guess-table-en}.

\begin{table}[h]
\centering
\caption{Top 15 guesses at each turn translated into English with accompanying the part-of-speech tags. Words that were the actual answers within these days are marked in bold.}
\begin{tabular}{|l|l|r|l|l|r|l|l|r|}
\hline
\multicolumn{1}{|c|}{\textbf{G1}} & \textbf{Tag} & \multicolumn{1}{c|}{\textbf{$\sum$}} & \multicolumn{1}{c|}{\textbf{G2}} & \textbf{Tag} & \multicolumn{1}{c|}{\textbf{$\sum$}} & \multicolumn{1}{c|}{\textbf{G3}} & \textbf{Tag} & \multicolumn{1}{c|}{\textbf{$\sum$}} \\ \hline
Sun & N nom sg & 5,341 & \textbf{Time} & N nom sg & 412 & \textbf{Time} & N nom sg & 433  \\ \hline
Wall & N nom sg & 3,179 & Sun & N nom sg & 355 & \textbf{Court} & N nom sg & 334 \\ \hline
\textbf{Court} & N nom sg & 1,579 & Day & N nom sg & 337 & \textbf{Fat} & N nom pl & 295 \\ \hline
Day & N nom sg & 1,476 & Linden & N nom sg & 290 & \textbf{Thing} & N acc sg & 273 \\ \hline
Luck & N nom sg & 1,449 & Cat & N nom sg & 284 & \textbf{Half} & N dat sg & 271 \\ \hline
Milk & N nom sg & 1,237 & Milk & N nom sg & 266 & \textbf{Time} & N acc sg & 247 \\ \hline
Bread & N nom sg & 1,217 & Apple & N nom sg & 262 & \textbf{Product} & N nom sg & 230 \\ \hline
Linden & N nom sg & 1,159 & Sun & N loc sg & 207 & \textbf{God} & N nom sg & 225 \\ \hline
Link & N nom sg & 958 & Wall & N nom sg & 205 & \textbf{Put} & V ptcp pst m & 224 \\ \hline
Box & N nom sg & 952 & Thing & N nom sg & 204 & \textbf{Halves} & N nom pl & 214 \\ \hline
Apple & N nom sg & 950 & Bread & N nom sg & 203 & \textbf{Market} & N acc sg & 214 \\ \hline
Cat & N nom sg & 869 & Tire & N nom sg & 192 & \textbf{Outfit} & N acc sg & 206 \\ \hline
Streets & N nom pl & 676 & Luck & N nom sg & 188 & \textbf{Peace} & N acc sg & 201 \\ \hline
School & N nom sg & 673 & \textbf{Fat} & N nom pl & 175 & \textbf{At once} & Adv & 200 \\ \hline
\hline
\multicolumn{1}{|c|}{\textbf{G4}} & \textbf{Tag} & \multicolumn{1}{c|}{\textbf{$\sum$}} & \multicolumn{1}{c|}{\textbf{G5}} & \textbf{Tag} & \multicolumn{1}{c|}{\textbf{$\sum$}} & \multicolumn{1}{c|}{\textbf{G6}} & \textbf{Tag} & \multicolumn{1}{c|}{\textbf{$\sum$}} \\ \hline
\textbf{Outfit} & N acc sg & 432 & \textbf{Clean} & V inf & 382 & \textbf{Tiles} & N acc pl & 345 \\ \hline
\textbf{Clean} & V inf & 382 & \textbf{Do} & V prs 2 pl & 364 & \textbf{Rainis} & N nom sg & 296 \\ \hline
\textbf{Half} & N acc sg & 382 & \textbf{Longer} & Adv cmp & 359 & \textbf{Pleasure} & N acc sg & 295 \\ \hline
\textbf{Later} & Adv cmp & 380 & \textbf{Basket} & N loc sg & 350 & \textbf{Gently} & Adv & 289 \\ \hline
\textbf{Away} & Adv & 364 & \textbf{Own} & Pron dat pl f & 340 & \textbf{Frequent} & Adj nom sg & 278 \\ \hline
\textbf{Time} & N nom sg & 354 & \textbf{Outfit} & N acc sg & 337 & \textbf{Built} & V ptcp pst f & 258 \\ \hline
\textbf{Course} & N loc sg & 353 & \textbf{Worth} & Adj N sg m & 334 & \textbf{Own} & Pron dat pl f & 250 \\ \hline
\textbf{Paint} & N acc sg & 340 & \textbf{Land} & N dat pl & 327 & \textbf{Take} & V deb & 245 \\ \hline
\textbf{Put} & V ptcp pst m & 339 & \textbf{Pour} & V prs 2 sg & 324 & \textbf{Wide} & Adj nom f & 243 \\ \hline
\textbf{Product} & N nom sg & 338 & \textbf{Away} & Adv & 322 & \textbf{Good} & Adj nom pl f & 238 \\ \hline
\textbf{Table} & N acc sg & 335 & \textbf{Good} & Adj nom pl & 321 & \textbf{Land} & N dat pl & 237 \\ \hline
\textbf{God} & N nom sg & 332 & \textbf{Later} & Adv & 321 & \textbf{Knowing} & V ptcp & 236 \\ \hline
\textbf{Block} & N acc sg & 331 & \textbf{Likewise} & Adv & 317 & \textbf{Pour} & V prs 2 sg & 234 \\ \hline
\textbf{Market} & N acc sg & 328 & \textbf{Take} & V deb & 317 & \textbf{Neck} & N acc sg & 226 \\ \hline

\end{tabular}
\label{tab:guess-table-en}
\end{table}


\newpage
\section*{Appendix C. Distributions of Words in the Corpora}

\begin{figure}[h]
    \centering
    \includegraphics[width=8.5cm]{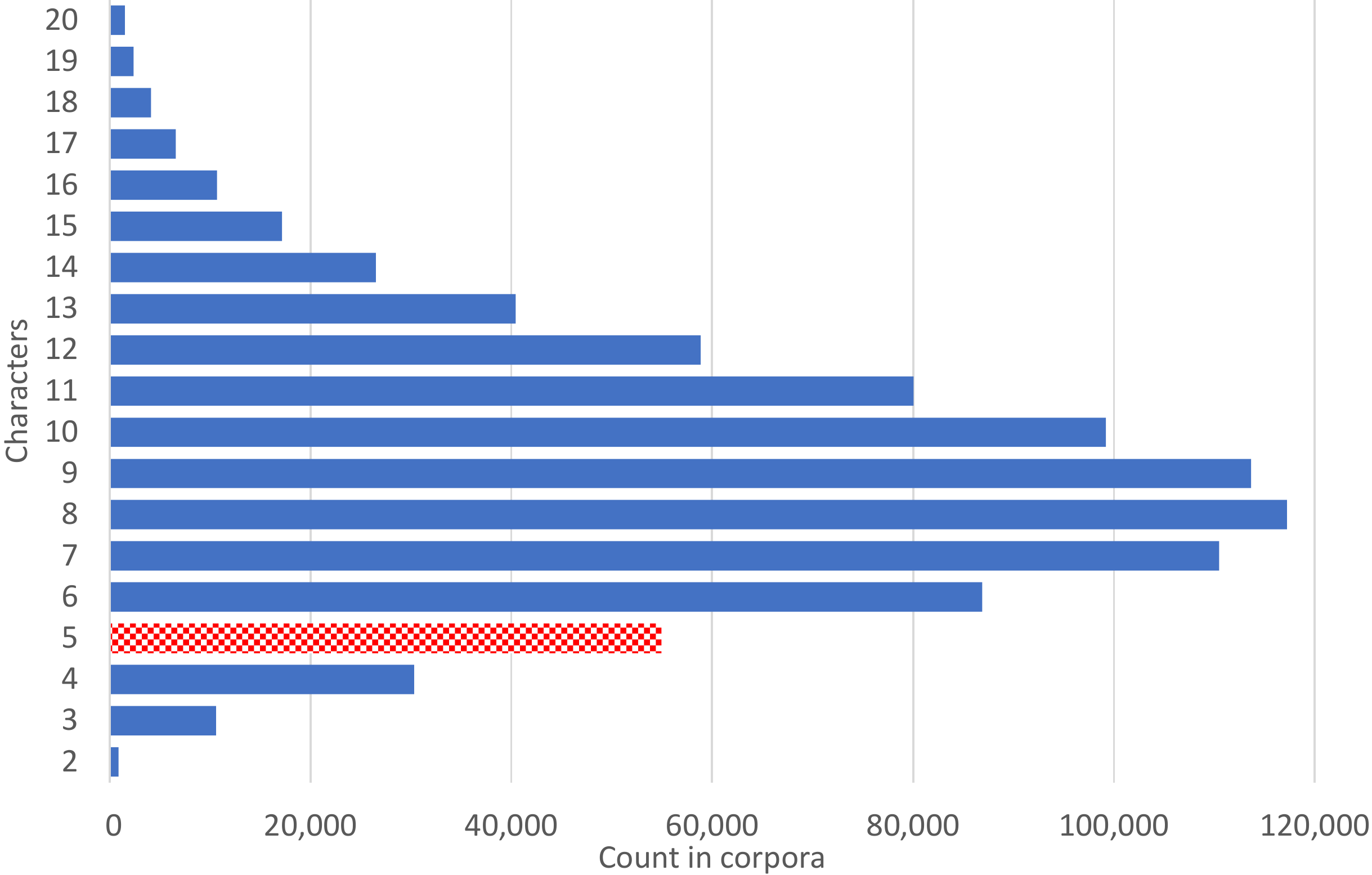}
    \caption{Distribution of unique n-character words in the corpora.}
    \label{fig:uniq-dist}
\end{figure}

\begin{figure}[h]
    \centering
    \includegraphics[width=8.5cm]{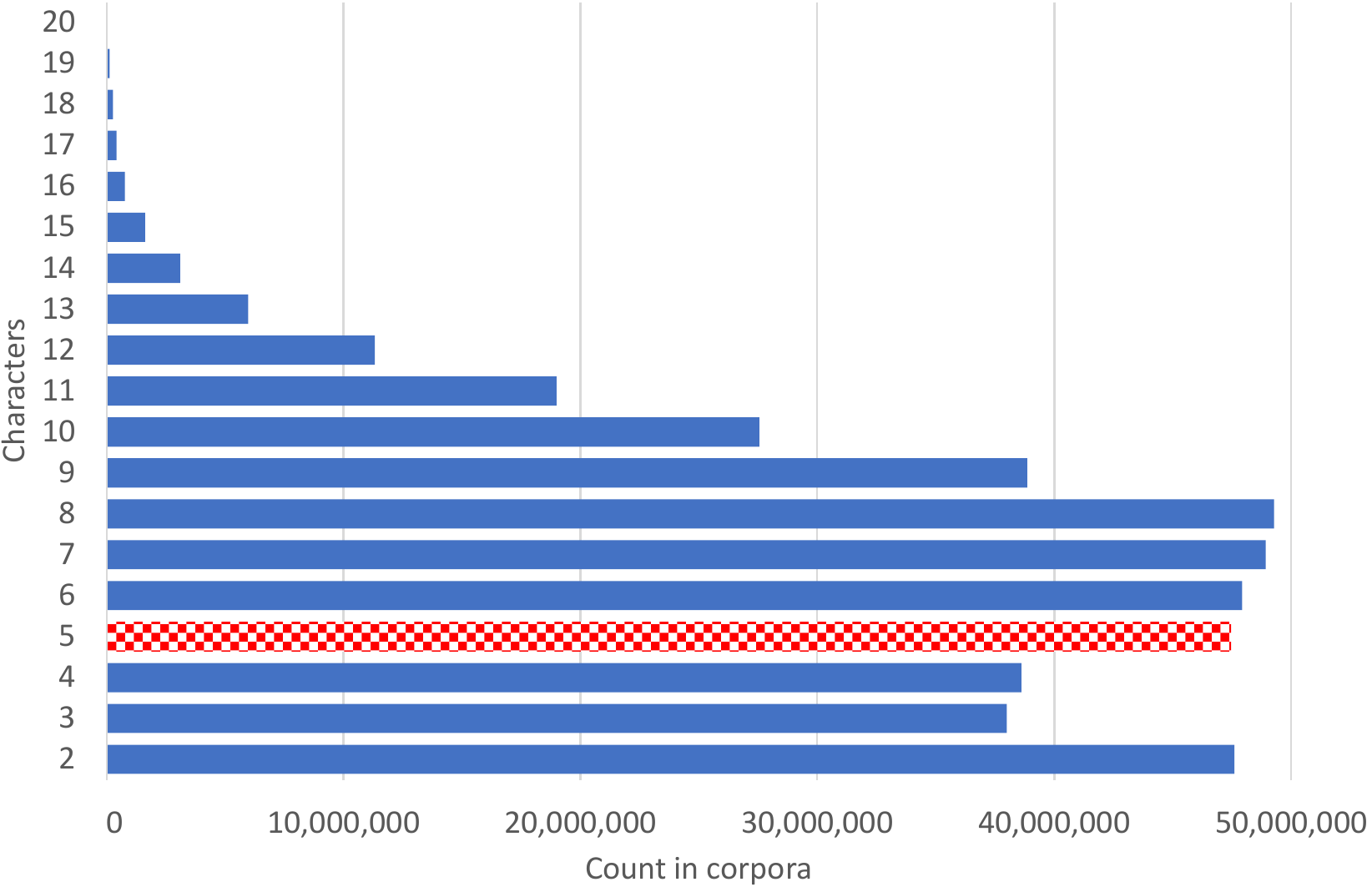}
    \caption{Total distribution of n-character words in the corpora.}
    \label{fig:total-dist}
\end{figure}

Figures \ref{fig:uniq-dist} and \ref{fig:total-dist} show the distribution of unique n-character words and total word counts of each length in the corpora of $\sim$32M unique Latvian sentences from Opus. We can see that 5-character words rank only 8th within the corpora, having 54,953 unique forms. However, in terms of total appearances in the corpora, 5-character words are almost as frequent as 2-character words, ranking 5th overall with 47,468,991 total appearances.



\received{August 19, 2022}{*}{August 22, 2022}
\end{document}